\documentclass[a4paper, 10 pt, conference]{IEEEtran}
\usepackage[left=1.9cm, right=1.3cm, bottom=2.1cm, top=3.66cm]{geometry}

\usepackage{eurosym}
\usepackage{upgreek}

\usepackage{cite}
\ifCLASSINFOpdf
  \usepackage[pdftex]{graphicx}
  \usepackage{import}
\else
\fi
\usepackage[english]{babel}
\usepackage[utf8]{inputenc}
\usepackage[]{caption}
\usepackage{subcaption}
\usepackage{placeins}
\usepackage{color}
\usepackage{mathtools}
\usepackage{placeins}
\usepackage{relsize}
\usepackage{amssymb}
\usepackage{units}
\usepackage{amsmath}
\usepackage{float} 		
\usepackage{tikz}

\usepackage{tikz}
\usepackage[color=black,opacity=1,angle=0,scale=1]{background}
\backgroundsetup{
contents={
\begin{tikzpicture}
\node at (current page.center) [align=center] {\textcopyright 2019 IEEE. Personal use of this material is permitted. 
\\ Permission from IEEE must be obtained for all other uses, in any current or future media, including reprinting/republishing this material for advertising or promotional purposes, 
\\ creating new collective works, for resale or redistribution to servers or lists, or reuse of any copyrighted component of this work in other works.
\\ DOI: 10.1109/ITSC.2019.8917240};
\end{tikzpicture}},
placement=bottom,
scale=0.6,
vshift=20
}

\begin{document}
\title{\vspace{0.5cm} 
Comfortable Priority Handling with Predictive Velocity Optimization for Intersection Crossings}

\author{\IEEEauthorblockN{Tim Puphal$^{1\dagger}$, Malte Probst$^{1\dagger}$, Misa Komuro$^{2}$, Yiyang Li$^{2}$ and Julian Eggert$^{1}$}
\IEEEauthorblockA{$^{1}$ Honda Research Institute (HRI) Europe, 
	Carl-Legien-Str. 30, 63073 Offenbach, Germany \\
	Email: {\tt\small \{tim.puphal, malte.probst, julian.eggert\}@honda-ri.de} \\
	$^{2}$ Honda Innovation Lab (HIL) Tokyo, Honda R\&D Co., Ltd. 9-7-1 Akasaka, 107-6238 Tokyo, Japan \\
	Email: {\tt\small \{Misa\_Komuro, Yiyang\_Li\}@n.t.rd.honda.co.jp} \\
	$\dagger$ The authors contributed equally to this work}}


%Telephone: (800) 555--1212, Fax: (888) 555--1212}

% use for special paper notices--

\maketitle

% maximal length 8 pages
\begin{abstract} %comfortable motion planning (i.e. front-before-back), (e.g. right-before-left)
We address the problem of motion planning for four-way intersection crossings with right-of-ways. Road safety typically assigns liability to the follower in rear-end collisions and to the approaching vehicle required to yield in side crashes. %An intersection manager has to define the order of crossing vehicles depending on the effective right-of-way. 
As an alternative to previous models based on heuristic state machines, we propose a planning framework which changes the prediction model of other cars (e.g. their prototypical accelerations and decelerations) depending on the given longitudinal or lateral priority rules. Combined with a state-of-the-art trajectory optimization approach ROPT (Risk Optimization Method) this allows to find ego velocity profiles minimizing risks from curves and all involved vehicles while maximizing utility (needed time to arrive at a goal) and comfort (change and duration of acceleration) under the presence of regulatory conditions. Analytical and statistical evaluations show that our method is able to follow right-of-ways for a wide range of other vehicle behaviors and path geometries. Even when the other cars drive in a non-priority-compliant way, ROPT achieves good risk-comfort tradeoffs.

\end{abstract}

% \fi
\IEEEpeerreviewmaketitle

\vspace{0.2cm}
\section{Introduction}
At intersections (e.g. Y, T and X junctions as well as roundabouts) and highway mergings (entering plus leaving ramps and overtaking), the driving task is simplified with traffic codes for prioritization \cite{trafficcode2000}. For cars, there are right-of-ways, stop lines or traffic lights. Pedestrians and bicycles utilize crosswalks and have general priority over cars. Even in simpler longitudinal scenarios, traffic participants should follow the direction of travel, keep on one side for multi-track streets and obey speed limits. These regulatory risks not only define who goes first, but also constrain the agents in their choice of actions and thereby make driving safer. Intentions become transparent and accessable for other agents so that they can be considered for motion planning.

Previous work models rules with state machines \cite{buehler2009}, ordering suitable maneuvers (keep distance, drive inside, etc.) based on the current kinematics of vehicle pairs. When entities do not comply to the norms, fallback plans avoid possible deadlock situations. Particularly for crossroads, the system should leave way or come to a stop when an obstacle takes precedence. 
This works reliably for normal driving, in critical conditions however it may generate reactive solutions that solely center on safety, effectively neglecting efficiency for the traffic flow. To robustly deal with a variable interplay of cars, it is therefore better to control behaviors dynamically using prediction-evaluation cycles that directly incorporate priority in the behavior finding procedure. %we present a priority velocity, efficiency and comfort 
%optimization

Our Risk Optimization Method (ROPT) presented in \cite{puphal2018} is an uncertainty-aware velocity planner which balances future integral risk from collisions and road structure with comfort and utility of the travel. In this paper, we expand its functionality to handle regulatory risk in vehicle-to-vehicle interactions (i.e., front, back, right and left geometric relations). Depending on the arising priority, ROPT first alters assumed trajectories as well as discounts awareness horizons for the respective other cars. %As an example, obstacles for which the ego car has precedence are considered only short times into the future and extrapolated to decelerate. 
%Segment heights and lags of multiple velocity snakes are then fine-tuned.
Next, fine-tuning slope and lag from segment-wise linear velocity profiles results into smooth ego car responses. We show in large-scale simulations that ROPT hereby successfully approaches, crosses and leaves uncontrolled intersections with varying behaviors (including cases where priority is violated) and taken paths of encountered traffic participants. %\ref{sec:outl} (including cases where priority compliance is violated), several

Section \ref{sec:rel} summarizes longitudinal, lateral and cooperative planning techniques in state-of-the-art. The description of a general multi-agent optimization framework is given in Section \ref{sec:fram} with emphasis on risk and comfort modeling. %into trajectory generation with Section \ref{sec:trajgen} and cost evaluation with Section \ref{sec:trajeval}. 
Sections \ref{sec:orderas} and \ref{sec:predprio} continue with priority assignment and prediction under regulatory risks. Finally, the analytical plus statistical experiments and evaluations are outlined in Section \ref{sec:exp} and Section \ref{sec:outl} presents our conclusion and prospects for future research. %for dynamic following as well as crossroads
%We show in large-scale simulations that ROPT thus finds 

\subsection{Related Work}
\label{sec:rel}
Vehicle control along same or parallel lanes is well established in the automotive industry. Here, the focus especially shifts from collision-free to likewise beneficial plans. Exemplarily in platooning \cite{geiger2012}, the minimized cost functional comprises the distance to all front vehicles for stable following. Traffic light assists \cite{treiber2014} create fuel-saving traverses during experienced phase switches (green, orange and red). When taking curves, \cite{liebner2013} apply a proactive deceleration with driver models and the course of lane changes are interpolated using Bezier curves in \cite{qian2016}. 

For crossing lanes with varying angles, possible driver intentions and ways of interaction become extensive. %\cite{galceran2015}.  as interactions  and their combinations
As a result on an intersection, right-of-way matrices \cite{erdmann2011} typically fill each lane relation with passing orders  %Finding the path conflicts and traversing distance to collision points, allows to define right-of-way matrices \cite{erdmann2011}. 
and safe maneuvers are then coordinated between the actors via if-then transition of defined driving states \cite{kammel2008}. While doing so, a fuzzy system \cite{lee1999} could dynamically alter single entries of the priorities, e.g. on behalf of emergency vehicles. In contrast to conventional heuristics, \cite{gregoire2014} also employ priority graphs to construct continous trajectories with safe gaps and \cite{plessen2016} iterate through priority schemes for realising orders even when each vehicle has to yield to another vehicle. 

\begin{figure}[t!]
      \centering      
      \resizebox{\linewidth}{!}{
      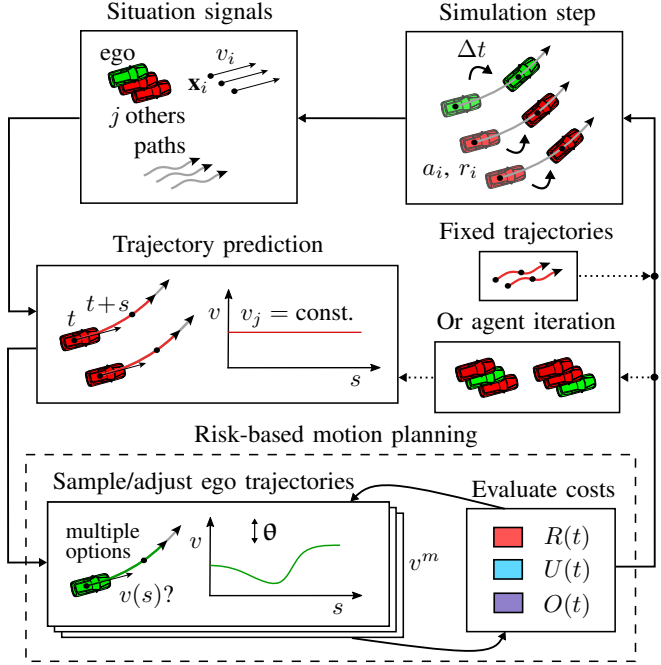}
      \caption{Concept of predictive velocity optimization. Assessing situation costs follows adapting multiple ego velocity profiles in repetitive cycles.} %block diagram
      \label{fig:simulator}
\end{figure}

Alongside lateral planning, recent research involves cooperative planning which considers optimized plans of other cars to locate global solutions. In \cite{frese2011}, priority-based approaches were evaluated as most efficient, but they cannot handle all scenarios. As a comparison, Monte Carlo tree search \cite{wolf2018} is used for lane merging with semantic vehicle-lane relationships and \cite{sezer2015} tested Markov decision processes under mixed observability for unsignalized intersections. Both methods lead to sensible ordering behaviors for specific complex scenarios which are implicitly influenced from the learned policy. %Merge or crossing orders are implictly given, because the global costs are otherwise elevated.
%solely

\vspace{0.2cm}
\section{Planning Framework}
\label{sec:fram}
\noindent %-block diagram with do path prediction, then trajectory generation with core and execute for ego
We tackle motion planning in structured environments by searching the velocity space $v$ over future times $s$. As depicted in Figure \ref{fig:simulator}, ROPT initially receives latest positions 
$\textbf{x}_i$, velocities $v_i$ and given map paths for the green ego car and $N_o$ other red cars (subscripted by $j$).\footnote{A traffic situation consists in this way of $N_o+1$ participants indexed with $i$.} Without prior knowledge, 
other trajectories are predicted on their respective paths with constant velocity up to a prediction horizon $s_h$. The goal of ROPT is now to optimize parameters $\boldsymbol{\uptheta}$ from
multiple velocity profiles $v^m$ for the ego agent. For this purpose, we alternate between adjusting $\boldsymbol{\uptheta}$ and evaluating risks $R(t)$, utility $U(t)$ and comfort $O(t)$ 
of the arising dynamic scene for the current time $t$. Once a defined cost threshold is satisfied for each sample, $v^m$ with the lowest cost is chosen and executed within a simulation step $\Delta t$ to obtain accelerations $a_i$ 
and jerks $r_i$. %actually when difference between function value or input value is under threshold %\footnote{We choose a plain double integrator, however vehicle dynamic models, such as the two-track model, can be added here if necessary.} %\footnote{Vehicle dynamic models, such as the two-track model, can be added if necessary. For our purposes, we choose however a plain double integrator.} 
In doing so, the simulator either updates other vehicles from measured fixed trajectories or controls them with their own planners. %Vehicle dynamic models, such as the two-track model, can be hereby added if necessary. For our purposes, we choose however a plain double integrator. %For the simulation, we use a simple double integrator. 

\subsection{Trajectory Optimization} 
\label{sec:trajgen}

\begin{figure}[t!]
      \centering
      \resizebox{\linewidth}{!}{
      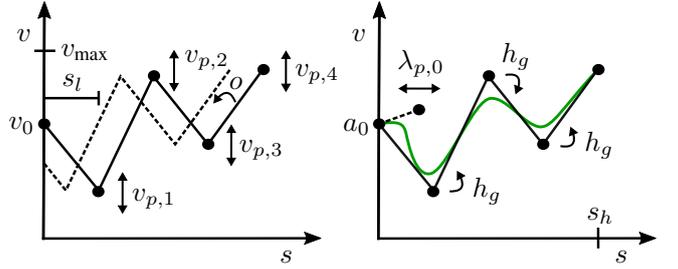}
      \caption{Left: Parameters and shift of chosen velocity snake. Right: Lag implementation and corner smoothing.} %exemplary
      \label{fig:snakes}
\end{figure}

%-Snake Trajectory \\
In complex scenarios with more than one risk source (i.e., driving in curve while crossing crowded intersection), the cost functional is non-convex. To overcome local minima, velocity shapes with high degrees of freedom are necessary. We choose for ROPT $n=4$ segments having fixed length $s_l=\unit[2.5]{\text{sec}}$ but variable end velocities $v_{p,n}$ (see left-hand side of Figure \ref{fig:snakes}, whereby $p$ stands for one parameter in the parameter set $\boldsymbol{\uptheta}$). %The first ramp starting the current velocity $v_0$ 
This allows to proactively plan tactical maneuvers, such as consecutively braking, keeping velocity and accelerating back. After each step $\Delta t$, the resulting ``snake'' profile is then time-shifted by an offset $o$ to match the new start velocity $v_0$ with same slopes $v_{p,n}$ for faster convergence. %After finding an optimal trajectory, $v_{p,n}$ can thus stay similar with less computational effort.        
Because $v(s)$ is discontinuous, we furthermore introduce an adjustable first lag $\lambda_{p,0}$ in the acting acceleration $a_0$. The right-hand side of Figure \ref{fig:snakes} shows that the following ramp transitions are supplementary smoothed with a Gaussian filter $h_g$.  %for reduced peaks

ROPT uses the non-gradient Powell's optimization method \cite{powell1964} which iteratively fits for $\boldsymbol{\uptheta}$ a quadratic function to three evaluation points and finds its vertex. %for each entry in theta %\footnote{On average, it takes less than 20 iterations to reach a suitable ego trajectory.} %A suitable ego trajectory is found usually in maximum 20 iterations. 
Soft constraints are set with penalizations for exceeding the minimal/maximal values $v_{\text{max}}$, $\lambda_{\text{min}}$, $a_{\text{min}}$ and $a_{\text{max}}$. Altogether, the optimization problem can thus be formulated as   
\begin{equation}
\text{min} \hspace{0.1cm} f \hspace{-0.05cm} \underbrace{(v_{p,1}, v_{p,2}, v_{p,3}, v_{p,4}, \lambda_{p,0})}_{\mbox{\footnotesize decision variables $\boldsymbol{\uptheta}$}} = \underbrace{R(t) - U(t) - O(t)}_{\mbox{\footnotesize fitness function $f$}},
%\text{min} f(\theta) = \underbrace{(\tau_0, v_1, v_2, v_3, v_4)}_{\mbox{\footnotesize decision variables}} = \underbrace{R(t) + O(t) - U(t)}_{\mbox{\footnotesize fitness function}} = C(t)
\label{eq: optimization}
\end{equation}
%\vspace{-0.1cm}
\begin{equation}
%\vspace{-0.4cm}
\text{subject to } v_{p,n} \leq  v_{\text{max}}, \hspace{0.15cm} \lambda_{p,0}  \geq \lambda_{\text{min}}, \hspace{0.15cm} a_{\text{min}} \leq  a_{p,n} \leq a_{\text{max}} \nonumber
%\text{subject to } \hspace{0.4cm} &\unit[0]{m/\text{sec}} \leq  v_i \leq  v_{\text{max}} \hspace{0.2cm} \nonumber \\ 
%&a_{\text{min}} \leq  a_i \leq a_{\text{max}} \\ %, \hspace{0.2cm} 
%&\unit[0]{\text{sec}} \leq  \tau_0 \leq \tau_{\text{max}} \nonumber
\label{eq: constraints}
\end{equation}
\noindent with segment accelerations $a_{p,n}$. A suitable ego maneuver is usually attained in less than 20 cycles. If not, we force the termination after a firm cycle number. %If not, we terminate the optimization after a fixed cycle number.  %For real-time capabilities, the optimization ends by default after 30 cycles. 
Besides the optimized snakes, we also sample fixed trajectories in our implementation: one constant velocity trajectory as well as one emergency stop and one acceleration trajectory. All trajectories are always evaluated in terms of their fitness, and one is in the end selected for behavior execution. Here, we introduce an hysteresis so that a switch to a different trajectory $v^m$ is done when the risk $R(t)$ of the new trajectory is relatively and absolutely smaller for a set period of time. 

\subsubsection{Smoothing Discrete Snake Function} 
%-picture with smoothing and tangent points and then next the smoothed verlauf of velocity \\
\noindent If we assume instantaneous actuation with fixed direct velocity points, ROPT may create trajectories which are unfeasible in real vehicles. The effective jerk $r(s)$ from $v(s)$ 
requires a continuous velocity curve. For this reason, we extrapolate the initial acceleration $a_0$ for the time $\lambda_{p,0}$ and blend its velocity line with the old ramp $(v_0,v_{p,1})$  according to   %and a weight factor $w(s)$
\begin{equation}
v(s) = v_0 +  (\lambda_{p,0}-s)a_0 + \frac{s}{s_l} (v_{p,1} - v_0)
\end{equation}
whereby $s=[0,\lambda_{p,0}]$. Afterwards, $v(s)$ is convoluted for the complete prediction interval $s_h$ with a Gaussian function %having constants $\sigma_s^2$ and $\mu_s$ given by
\begin{equation}
h_g(s)=N(\sigma_s^2,\mu_s=0). 
\end{equation}
We set the variance $\sigma_s^2$ and use $\mu_s=0$ to achieve further smoothness of the overall velocity curve without overshooting. As the derivatives $a(s)$ and $r(s)$ are numerically recalculated after the smoothing steps for $v(s)$, errors from asynchronicity are prevented. 

Consequently by optimizing $\lambda_{p,0}$, ROPT is able to influence the course of $r(s)$ (i.e., gradual actuation). Limits for $\lambda_{p,0}$ have to be enforced even when high-risk situations occur. The average brake lag to decelerate at once from $0$ to $a_{\text{min}}$ amounts to $\lambda_b=\unit[0.4]{\text{sec}}$ and engine acceleration to $a_{\text{max}}$ takes $\lambda_e = \unit[0.8]{\text{sec}}$.\footnote{In contrast, the action of taking the foot off the brake or gas pedal has immediate effect on the car.} With this in mind, we qualify the minimal lag threshold $\lambda_{\text{min}}$ depending on the acceleration $a_0$ as
\begin{equation}
\text{if } a_0 \geq 0 \text{: } \lambda_{\text{min}} = \frac{a_0}{a_{\text{max}}} \lambda_e , \hspace{0.2cm} \text{else: } \lambda_{\text{min}} = |\frac{a_0}{a_{\text{min}}}| \lambda_b. 
\end{equation}
%-$\tau_{\text{min}}=\unit[0.4]{\text{sec}}$ if $a<0$, $\tau_{\text{min}}=\unit[0.8]{\text{sec}}$ if $a>0$
Compared to employing continuous polynoms, our modified snake behaves smoothly and does not require the solution of a linear equation system to map $\boldsymbol{\uptheta}$ to the function shape.
In non-risky scenarios, ROPT is hence able to concentrate on comfortable behaviors. %has the advantage that there is no need of solving a linear equation system.

\subsection{Risk, Utility and Comfort Prediction}
\label{sec:trajeval}
\noindent
In the following, we look at one future plan $v^m(s)$ for the ego vehicle combined with constant velocities $v_j(s)$ of the other vehicles. %The evaluation step treats the contained dynamic 
This subsection describes the computation of the entire accumulated future costs $R(t)$, $U(t)$ and $O(t)$ contained in the resulting scene state sequence $\textbf{z}_{t:t+s}$, starting from time $t$ and evolving over $s$.  %\ref{sec:trajeval}

For the risk analysis, we can only postulate that $\textbf{z}_{t:t+s}$ will happen with a certain probability (e.g. because of sensor inaccuracies or unkown drivers' intention). % from collision and curve
%ROPT combines two probabilistic methods a Gaussian method for an instantaneous collision probability with the survival analysis
%[6] to retrieve an accumulated critical event probability. 
On this account, ROPT models accident occurences within an inhomogeneous Poisson process. The total event rate $\tau^{-1}(\textbf{z}_{t+s})$ describes the \makebox[\linewidth][s]{mean time between events. When subdivided into an escape}  \par
\noindent rate $\tau_0$ (behavioral options mitigating dangers) and critical rates of collisions $\tau^{-1}_{\text{crit},j}$ as well as of losing control in curves $\tau^{-1}_{\text{curv}}$, we gain %\footnote{To compute $\tau_{\text{coll}}$, we take the overlap of 2D normal distributions around the expected positions between car pairs and $\tau_{\text{coll}}$ compares }
\begin{equation}
\tau^{-1}(\textbf{z}_{t+s}) = \tau_0 + \sum_j  \tau^{-1}_{\text{coll,j}}+\tau^{-1}_{\text{curv}}.
\vspace{-0.04cm}
\label{eq:taucrit}
\end{equation} 
In Equation (\ref{eq:taucrit}), normal distributions are additionally defined for the positions and velocities growing after each prediction step size $\Delta s$. While $\tau^{-1}_{\text{coll}}$ is dependant on the distances $d_j(s)$ of car pairs, $\tau^{-1}_{\text{curv}}$ takes the lateral ego acceleration $a_{\text{y}}(s)$ into account.\footnote{For further details about the Gaussian method, please refer to \cite{puphal2018}.} 

Since the conveyed kinetic energy in a casualty is proportional to the operating masses $m_i$ and velocity vectors $\textbf{v}_i$, we use for collision and curve damage
\begin{equation}
D_{\text{coll},j}(s;t,\Delta s) = D_0+ \frac{m_1m_j}{2(m_1+m_j)}\| \textbf{v}_j - \textbf{v}_1 \|^2,
\end{equation}
\begin{equation}
D_{\text{curv}}(s;t, \Delta s) = D_0 + \frac{1}{2}m_1 \| \textbf{v}_1 \|^2
\end{equation}
with an offset $D_0$. Anytime a crash is not possible conditional to kinodynamics of the cars, $D_{\text{coll},j}$ is set to $0$.
%The kinetic energy of the accident (proportional to velocity vectors $v_i$) is 
By introducing a survival probability that the ego entity will not be engaged in an event during $[t, t+s]$ via
\begin{equation}
S(s;t,\textbf{z}_{t:t+s})=\exp\{-\int_o^s \tau^{-1}(\textbf{z}_{t+s'}) \, ds'\},
\end{equation}
we eventually obtain $R(t)$ as the temporal integration of rates, damages and probabilities%\footnote{A straightforward numerical calculation of the integral is sufficient with small $\Delta s$.}
\begin{equation}
R(t) = \int_0^{\infty} (\sum_j \tau_{\text{coll},j}^{-1}D_{\text{coll},j}+\tau_{\text{curv}}^{-1}D_{\text{curv}})S \,ds.
\label{eq: risk}
\end{equation} 
A straightforward numerical calculation of the integral is sufficient with small $\Delta s$, e.g. we utilize $\unit[0.05]{\text{sec}}$.

Besides minimizing risk, ROPT maximizes benefit (i.e., utility and comfort) as well. The considered utility consists of the overall needed time to arrive at the goal affected from the ego velocity $v_1$ and deviations to the desired velocity $v_d$. The components are weighted with driver-specific constants $b^t$ and $b^d$ to retrieve
\begin{equation}
U(t) = \int_0^{\infty} (b^t |v_1| + b^d |v_1 - v_d|) S \,ds.
\label{eq: utility}
\end{equation} 
Comfort returns are granted if the behavior does not change (ego acceleration $a_1\approx 0$) and the approach to planned $a_1$ is slow (ego jerk $j_1\approx 0$) so that 
\begin{equation}
O(t) = \int_0^{\infty} -(b^c |a_1| + b^j |j_1|) S \,ds.
\label{eq: comfort}
\end{equation} 
Calibrating the occuring parameters $b^c$ and $b^j$ in combination with $b^t$ plus $b^d$ allows to reproduce different driver characteristics, such as conservative versus sporty. The costs are therefore expressed in the same unit \euro \hspace{0.04cm}. For higher $s$, we also consider the survival function $S$ in Equation (\ref{eq: utility}) and (\ref{eq: comfort}) so that predicted benefits cannot surpass risk factors. % High-risk situations should not be surpassed for benefit reasons.

\section{Regulatory Risks}
\label{sec:regrisk}
At least one car is generally responsible in an accident \cite{shalev2017}. For example, during a following scenario the back vehicle is hold liable, if it failed to keep safe distances to the leading vehicle. %required to always keep safe distances to the front vehicle. Proper responses can thus be applied when the leading behavior changes. If the follower fails to do so, the responsibility for the crash lies with him. 
%
%\noindent Shalev-Shwartz \\
In contrast, for car pairs frontally driving against each other, both are seen at fault. Then, priorities around intersections with traversing paths allow to shift the responsibility on the driver who had to yield. A requirement for these longitudinal and lateral circumstances is that the superior entity (leading or prioritised car) did not brake or accelerate unreasonably. Otherwise in law, the share of the blame and costs is again divided among the involved parties.
%An important research question for autonomous vehicles is how to mathematically implement this responsibility system.

%In previous research, we constructed, for the ego car
To implement asymmetry in interactions, we formerly treated situations as discrete awareness or non-awareness entity combinations \cite{damerow2015}. By iterating over each and superposing the inherent risks, an optimal trajectory was constructed. However for ROPT, a more computationally efficient way is to only focus on the likely situations based on priorities. ROPT thus a) categorizes the path relation between vehicles plus matches them to legal right-of-ways (e.g. front-before-back, right-before-left) and b) modifies appropriately the behavior-relevant prediction model of other cars (i.e., altering the influence on own risk and calculating different trajectories). 

\vspace{0.15cm}
\subsection{Order Assignment}
\label{sec:orderas}
\begin{figure}[t!]
      \centering
      \resizebox{\linewidth}{!}{
      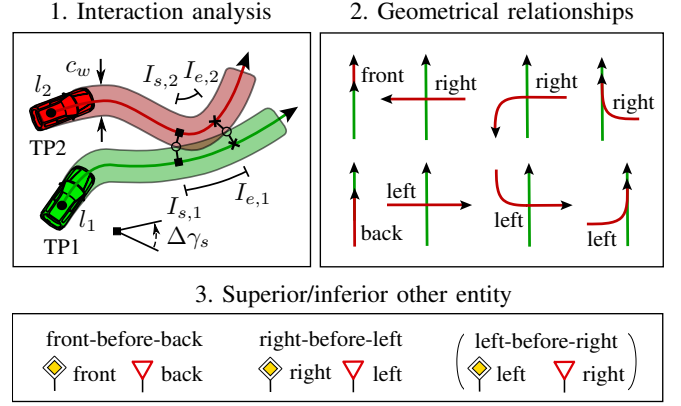}
      %\caption{Left: Collision risk prediction with survival analysis. Right: Interaction parameters.}
      \caption{Individual steps for regulatory risk estimation on the basis of spatial path corridors.} 
      \vspace{0.1cm}
      \label{fig:interaction}
\end{figure}

A generic driving scene of two traffic participants (TP) with $i=1,2$ is illustrated in Figure $\ref{fig:interaction}$. As a starting point, we trail corridors having widths $c_w$ from their current longitudinal position $l_1$ and $l_2$ until the trajectory end. Subsequently, the zone of interaction is given where both corridors interfere. We project start and end points to each path and get separate boundaries $I_{s,1}$, $I_{e,1}$ for TP1 and $I_{s,2}$ and $I_{e,2}$ for TP2. %\footnote{From now on, the second subscript denotes the participant number $i$ and not velocity segment order $n$.}
 
In the longitudinal case, one or both TP's are in the interaction zone at moment $t$. Comparing the positions $l_i$ allows to assign TP2 being in front or in the back to TP1. In total, we can write
\begin{equation}
l_1 \in [I_{s,1}, I_{e,1}] \wedge l_2 \in [I_{s,2}, I_{e,2}] \rightarrow
\end{equation}
\begin{equation}
\text{front: } l_1 < l_2 \text{, back: } l_1 > l_2. \nonumber
\end{equation}
For the lateral case, the trajectories meet in the future. When we look at the difference angle $\Delta \gamma$ of the interaction start $I_{s,1}$ and $I_{s,2}$, TP2 is to the right or left depending on its value in compliance with
%\noindent -Lateral case: trajectories intersect/merge in prediction horizon \\ 
%-depending on difference angle at start points of interaction zone, other car left or right
\begin{equation}
\angle I_{s,1} I_{s,2} = \gamma_{s,1} - \gamma_{s,2} = \Delta \gamma_s,
%\Delta \gamma_b = \gamma_{b,1} - \gamma_{b,2} 
\end{equation}
\begin{equation}
\text{right: } \Delta \gamma_s \in (0, \pi) \text{, left: } \Delta \gamma_s \in (\pi, 2\pi).
%\angle l_{b,1} l_{b,2} = \Delta \gamma_b
%\Delta \gamma_b = \gamma_{b,1} - \gamma_{b,2} 
\end{equation}

Possible interaction types for TP1 driving fixed from the bottom to the top on X-intersections are also summarized in Figure $\ref{fig:interaction}$. Besides TP's driving on the same path, the trajectory of TP2 can intersect, be curved before or after and merge with trajectory of TP1. For front-before-back, TP2 is superior in front and inferior in back relations. Analogously, right-before-left determines TP2 as superior for right and inferior for left contexts. In other countries with left-before-right, the order assignment is switched.

\subsection{Prediction under Priority}
\label{sec:predprio}
\begin{figure}[t!]
      \centering
      %\resizebox{\linewidth}{!}{   
      %\import{img/}{interaction_examples3.pdf_tex}}
      \resizebox{\linewidth}{!}{
      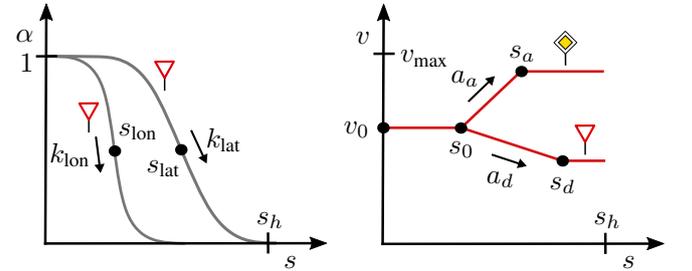}
      \caption{Left: Change in collision risk over future times. Right: Acceleration and deceleration assumptions for other entity. \vspace{-0.3cm}}
      \vspace{0.2cm}
      \label{fig:modpred}
\end{figure}
%\vspace{-0.15cm}
\subsubsection{Awareness Discounting} %Reduce prediction horizon (not entirely true, because we still predict comfort and utility with full prediction horizon)
On crowded public roads, we concentrate on the main cars around which have right-of-way. The remaining cars are solely considered if they come critically close.
In this sense, ROPT discounts the collision risk $\tau_{\text{coll},j}^{-1}$ of inferior obstacles with a monotonically decreasing function. Regarding longitudinal interactions, our
sigmoid function $\alpha_{\text{lon}}(s)$ is described with the slope $k_{\text{lon}}$ and midpoint $s_{\text{lon}}$ which leads to
%\vspace{-0.2cm}
\vspace{0.05cm}
%\noindent -when ego superior entity: discount influence of other car in collision risk with monotonically decreasing function, here logistic function \\
\begin{equation}
\alpha_{\text{lon}}(s) = 1 - \frac{1}{1+\text{exp}\{k_{\text{lon}}(s-s_{\text{lon}}) \}} \hspace{0.01cm},
\end{equation}
%\vspace{-0.5cm}
\vspace{0.01cm}
\begin{equation}
\tau_{\text{coll},j}^{* \hspace{0.04cm} -1}(\textbf{z}_{t:t+s}) = \alpha_{\text{lon},j}(s) \tau_{\text{coll},j}^{-1}(\textbf{z}_{t:t+s}). 
\label{eq:riskmod}
\end{equation}
\vspace{-0.3cm}

\noindent The equations for $\alpha_{\text{lat}}(s)$ are the same, whereby the chances that the other vehicle perceives us is lower in intersection scenarios and parameters $k_{\text{lat}}$ and $s_{\text{lat}}$ are set higher (compare Figure \ref{fig:modpred} on the left).
%-formulas for $\alpha_{\text{lat}}(s)$ are the same with parameters $k_{lat}$ and $s_{lat}$

%-chances that other car see me are larger in longitudinal following and thus the parameters can be set to lower values, whereby for intersection scenario it has to be higher prediction horizon \\
\subsubsection{Delayed Acceleration Patterns}
\noindent Without priority knowledge, vehicles are extrapolated with constant velocity from Section \ref{sec:fram}. In addition to decreasing the awareness, ROPT predicts delayed accelerations in the lateral situation as well.
If the other car is superior, the ego planner assumes first constant velocity $s_0$ long, an acceleration phase $(s_a, a_a)$ and ultimately steady velocity up to $s_h$. The case differentiation follows as
\begin{equation}
v(s)=\begin{cases}
  v_0,  & \text{for }s = [0, s_0), \\
  v_0 + a_a(s - s_0), & \text{for }s = [s_0, s_a], \\
  v_0 + a_a(s_a - s_0), & \text{for } s = (s_a,s_h].
\end{cases}
\label{eq:pattern}
\end{equation}
\vspace{-0.1cm}

\noindent Here, the strength of $a_a$ depends on the active velocity $v_0$ (i.e., we apply $a_a\hspace{-0.09cm}=\hspace{-0.09cm}0$ for $v_0\hspace{-0.09cm}=\hspace{-0.09cm}v_{\text{max}}$ and linear growth to $a_a\hspace{-0.04cm}=\hspace{-0.04cm}a_{\text{max}}$ when $v_0\hspace{-0.04cm}=\hspace{-0.04cm}0$). 
This is based on the fact that applied accelerations around intersections are statistically stronger from standstill. By comparison if the other car has superior relations, ROPT uses a longer deceleration phase $(s_d, a_d)$ with unchanged $a_d$. %The strength \footnote{We add that when the car has $v_j\approx0$, it accelerates more than when $v_j \gg 0$.} 

In a last step, we clip the velocities $v_{j}(s)$ to be higher than $0$ and lower than the maximal curve velocity $v_{c,j}(s)$ and allowed limit $v_{\text{max}}$ with
\vspace{-0.15cm}
\begin{equation}
v_j(s) = \text{max}(v_j(s), 0),
\end{equation}
\begin{equation}
v_j(s) = \text{min}(v_j(s), v_{\text{c},j}(s), v_{\text{max}}). 
\end{equation}
 %\rightarrow v_{max} road constraint
The altered velocity patterns from Figure \ref{fig:modpred} on the right lead to better predictions of other vehicles, given that they behave according to the traffic rules. Because of the delay $s_0$ in combination with the start of unawareness $s_{\text{lon}}$ and $s_{\text{lat}}$, ROPT is in short-times even robust against moderately wrong assumptions. %(i.e. does not obey to the traffic rule). 
Each should be set that no crash happens for any acceleration or deceleration maneuver. More detailed predictions can be achieved by considering environment conditions (just accelerating in interaction zone and coming to halt at stop line), participant types (e.g. motorbike or truck) and occuring situation class (highway versus inner city). 

%reference powell, th and pet
\vspace{0.38cm}
\section{Experiments}
\label{sec:exp}
We want to show in our simulations that ROPT can handle a wide array of interactions which typically occur at intersection crossings and that the planned solution is compliant with priority rules.   %As it turns out, the regulatory risks lead to an asymmetry in the solutions of front vs. back and left vs. right while keeping safety and comfort at desirable levels. 
For this reason, we first analyze in Section \ref{sec:analyticalvar} one vehicle pair during dynamic followings before or after crossroads as well as during passing behaviors within intersection areas.  %  longitudinal plus intersection scenarios where the actions of the other car are varied. The longitudinal scenario represents thereby dynamic following as it occurs predominantly before and after intersections, whereas the intersection scenario is particularly suited to investigate passing behavior. 
We hereby show quantitatively the effect of the altered prediction models from ROPT. Second in Section \ref{sec:randomvar}, we randomize the possible paths for the two cars in test statistics to establish the robustness of ROPT in terms of risk and comfort. As it turns out, the optimization compensates non-priority-compliant other behavior with adequately elevated jerks.  
\subsection{Analytical Variation of Other Behavior}
\label{sec:analyticalvar} % the two
\begin{figure}[t!]
      \centering
      \resizebox{\linewidth}{!}{
      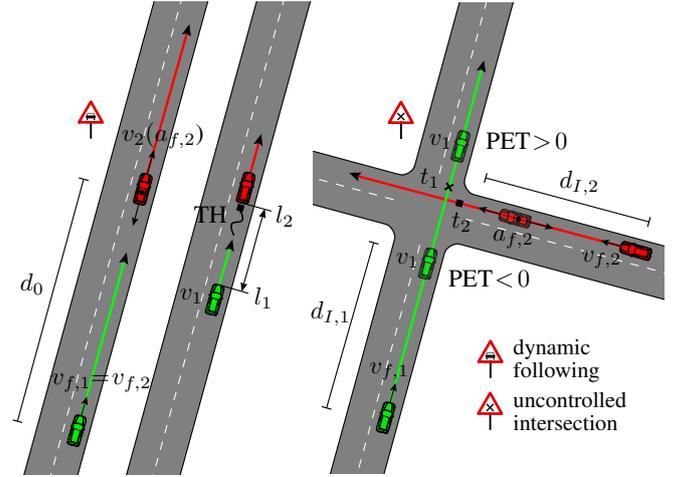}
      \caption{Left: Initial plus final conditions in a following scenario under front-before-back priority (case ego following). Right: Two possible scene evolutions for an intersection with right-before-left (case other priority).} % (case ego following), (case other priority) The ego car (green) is controlled by ROPT and represents TP1. Depicted in red, the competing other car is TP2.} %the competing other car (red) represents TP2.} %and the competing other car (red) represents TP2.} %}%\footnotemark} 
      %The ego car (green) is controlled by ROPT and represents TP1. Depicted in red, the competing other car is TP2.
      \vspace{-0.05cm}%} %}
      \label{fig:genericsituations}
\end{figure}

Both regarded basic scenarios are pictured in Figure \ref{fig:genericsituations}: longitudinally driving behind a leading TP to the front and an uncontrolled intersection having a second TP to the right. We also reproduce the cases that TP2 is in the back or approaching from the left. In each case, we vary for TP2 the initial velocity $v_{f,2}$ in between $0$ and $\unit[15]{m/\text{sec}}$. %\footnote{From now on, the second subscript in $v$ denotes the participant number $i=1,2$ and not snake segment $n=1\small{-}\hspace{0.05cm}4$.}
After $\unit[1]{\text{sec}}$, a deceleration/acceleration $a_{f,2}$ is applied in the range from $-3$ to $\unit[3]{m/\text{sec}^2}$ for the duration of $\unit[3]{\text{sec}}$. The challenge for ROPT is then to adapt TP1 (ego car, green) to the fixed actions of TP2 (other car, red) 
%to the fixed actions of TP2 
while considering the regulations front-before-back and right-before-left. Concerning the longitudinal environment, ROPT starts at a distance $d_0\hspace{-0.03cm} = \hspace{-0.03cm} \unit[50]{m}$ to TP2 and with equal speed $v_{f,1}\hspace{-0.05cm}=\hspace{-0.05cm}v_{f,2}$. A soft road limit of $v_{\text{max}} = \unit[20]{m/\text{sec}}$ is also valid. For the intersection instance, beginning offsets until the path corridors overlap are chosen as $d_{I,1}\hspace{-0.03cm}=\hspace{-0.03cm}d_{I,2}\hspace{-0.03cm}=\hspace{-0.03cm}\unit[40]{m}$ and the velocity parameters of ROPT (i.e., $v_{f,1}$ and desired velocity $v_{d,1}$) are set to $\unit[10]{m/\text{sec}}$.   %With these parameter settings, the 3-second deceleration process of the other car always finishes before the stop line of the intersection is reached.
%density of simulation runs is 0.5 for v and a

%control the velocity $v_1$ in TP1

%put variables for other car (v_0 and a) and visualize pet and th
% on top of plots other priority and ego priority, in caption other car to the right, left, etc. (or other superior, other inferior)

%\footnotetext{From now on, the second subscript in $v$ denotes the participant number $i$ and not segment order $n$.}

In the evaluation, we particularly look at the indicators Time Headway ($\text{TH}$) \cite{transpres2010} %eventually
and Post-Encroachment Time ($\text{PET}$) \cite{brian1978} which depend on the kinematics of the vehicles  
\begin{align}
\text{TH} = \frac{-\Delta l}{v_1} \ \text{with} \ \Delta l = l_1 - l_2, \label{eq:th} \\
\text{PET} = -\Delta t \ \text{with} \ \Delta t = t_1 - t_2. 
\end{align}
%  when the other car is in front of the ego car, for the inverted case the other car TH is used.} 
The events $t_1$ and $t_2$ indicate in $\text{PET}$ when the ego entity leaves and when the obstacle enters the interaction zone, respectively. On that account, ROPT can either pass in front with $\text{PET} >0$ or behind with $\text{PET}<0$. 
For $\text{TH}$, we extract the stable value $\text{TH}_{\text{stable}}$ once a constant longitudinal distance $\Delta l$ is maintained.\footnote{Equation (\ref{eq:th}) counts if the ego car follows another vehicle. For the inverted case, the indices in TH are swapped.} To complete the utility assessment of ROPT, we eventually capture the lower boundary $v_{\text{low},1}$ and upper boundary $v_{\text{up},1}$ from the executed velocity course $v_1$. 

\begin{figure}[t!]
      \centering
      \resizebox{\linewidth}{!}{
      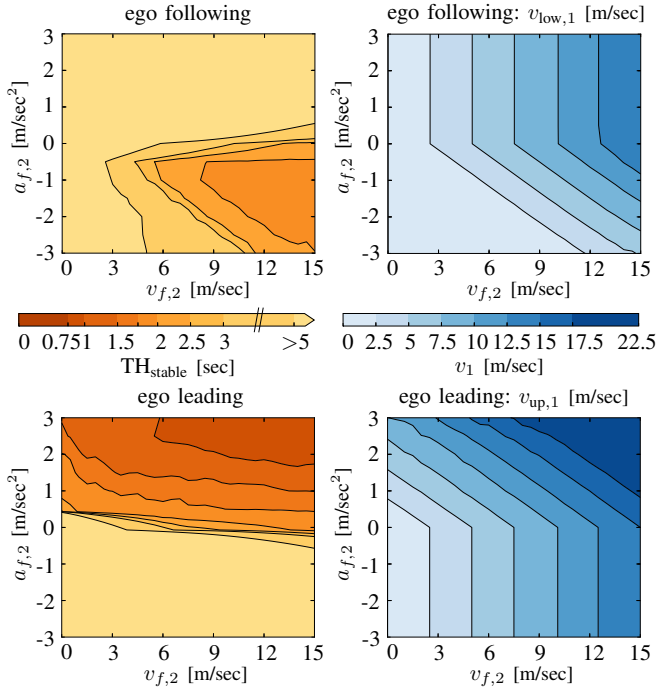}

      \caption{Indicators of ROPT behavior (minimal and maximal velocity) and its interplay with other car (i.e., stable headway) for range of fixed other actions (varying inital speed and acceleration). The priority-dependant awareness horizons lead to lower distances for back vehicles.}%higher distances for front vehicles.} 
      \vspace{-0.05cm}
      \label{fig:frontbeforeback}
\end{figure}

\subsubsection{Dynamic Following}
%First, explain what the Figures show, and how they have to be read (interpreted). Then, explain what ``good" behavior would look like (risk/utility/comfort) -> crucial that the reader understands this
An agent controlled with symmetric risk calculations for front and back would react very sensitively to following cars, e.g. in the case of tailgaiting. 
Due to the longitudinal risk discounting with Equation (\ref{eq:riskmod}), it is now harder for the back vehicle to push ROPT in front. The inferior entity is however not entirely ignored, since non-reaction can result in partial legal blame.  %motivate why it makes sense to not ignore the other vehicle in the back compleletly (braking abruptly can be dangerous and is forbidden /will lead to partial blame)
The contour plots within Figure \ref{fig:frontbeforeback} sort measured $\text{TH}_{\text{stable}}$ and extrema of $v_1$ into colored bins for every other trajectory point ($v_{f,2}$,~$a_{f,2}$).\footnote{The grid step size amounts to $\Delta_{x,y} = 0.5$ with linear interpolations.} %for the x- and y-direction
As can be seen at the bottom row, we allow small but sufficient $\text{TH}_{\text{stable}}$ until $\unit[1]{\text{sec}}$ to the back. If $a_{f,2}$ is positive, higher $v_{f,2}$ lead to decreasing $\text{TH}_{\text{stable}}$. At the same time, the final maximum velocity $v_{\text{up},1}$ matches the accelerating follower with $v_{f,2}\hspace{-0.01cm}+\hspace{-0.01cm}\unit[3]{\text{sec}}\cdot a_{f,\text{2}} \hspace{-0.03cm}<\hspace{-0.03cm}\unit[22.5]{\text{m/sec}}$. %As $v_{\text{max}}$ is a soft constraint, ROPT is allowed to apply higher values in $v_{1}$ once collision risk is immanent. %\footnote{For $v_{f,2}=$ and $a_{f,2}$} 
For negative $a_{f,2}$, ROPT is not influenced by the decelerating obstacle (i.e., $\text{TH}_{\text{stable}}$ from $\unit[3]{\text{sec}}$ upwards) and delivers steady velocity (i.e., $v_{\text{up},1}~\hspace{-0.13cm}=~\hspace{-0.13cm}v_{f,1}$ applies). %(i.e. $\text{TH}_{\text{stable}}>\unit[5]{\text{sec}}$)

In contrast, front vehicles with priority yield more proactive ego behaviors. When the leader brakes down, ROPT uses unaltered collision risks and converges to moderate $\text{TH}_{\text{stable}}\approx \unit[2]{\text{sec}}$ for large $v_{f,2}$ and $|a_{f,2}|$ (compare top row of Figure \ref{fig:frontbeforeback}). Is the other trajectory a stopping trajectory, the minimum end speed $v_{\text{low},1}$ becomes $\unit[0]{m/\text{sec}}$ in ROPT and thus $\text{TH}_{\text{stable}}$ exceeds $\unit[5]{\text{sec}}$. Unlimited $\text{TH}_{\text{stable}}$ are moreover also carried on, when TP2 is moving away with $a_{f,2}>0$. ROPT is therefore able to retain the varying beginning velocity, more specifically $v_{\text{low},1}=[\unit[0]{\text{m/sec}},\unit[15]{\text{m/sec}}]$.      

\subsubsection{Intersection Passing} %as mentioned before
\begin{figure}[t!]
      \centering
      \resizebox{\linewidth}{!}{
      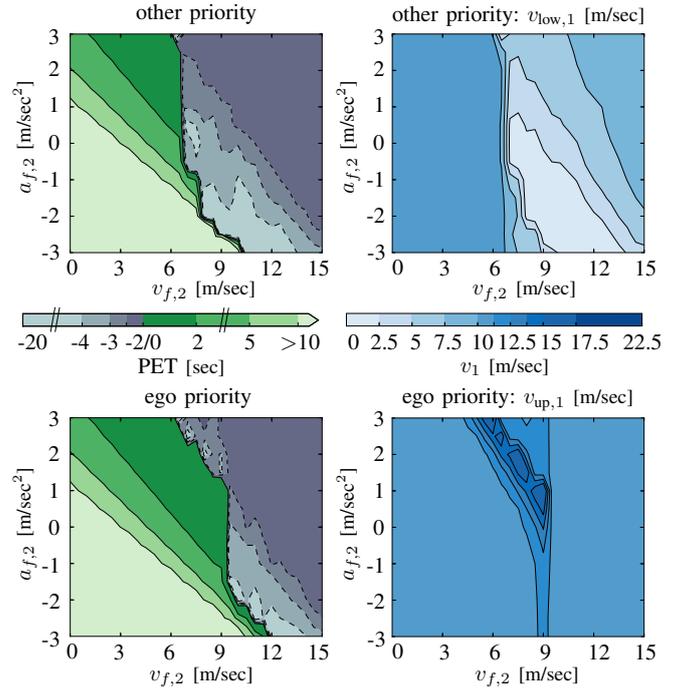} %
      \caption{Results for ROPT with altered velocity extrapolations. Top: Other car approaching from right. Bottom: Other car coming from left. On an intersection, ROPT accelerates more frequently when having priority.} %brakes more frequently for prioritized vehicles.}
      %\caption{Post-Encroachment Time in 2-car intersection scenario with right vs. left priority. Top row: Other car approaching from right. Bottom row: Other car approaching from left. On an intersection, ROPT either drives more often after the other prioritized car ($\text{PET}<0$) or takes its precedence ($\text{PET}>0$).} %with normal comfort ranges %The comfort is never disregarded (low $j_1$)
      \label{fig:leftbeforeright}
\end{figure}

Simple heuristical go/no-go decisions cannot ensure optimal driving cost tradeoffs for lateral priorities. Another entity on the right might be far away or decelerating so that the ego utility is neglected. More importantly, inferior cars that do not respect the right-of-way create arbitrary risk or discomfort peaks. 
Via the delayed acceleration patterns from Section \ref{sec:predprio}, ROPT is capable to continuously weigh benefits with risks for passing a rule-based intersection first or second. Figure \ref{fig:leftbeforeright} visualizes the isolines of $\text{PET}$, $v_{\text{low},1}$ or $v_{\text{up},1}$ for the same parameter variations in $v_{f,2}\hspace{-0.03cm}$ and $\hspace{-0.03cm}a_{f,2}$. While there is more area of $\text{PET}\hspace{-0.03cm}<\hspace{-0.07cm}\unit[-2]{\text{sec}}$ (i.e., ROPT driving second) when the car comes from the right, the condition $\text{PET}\hspace{-0.03cm}>\hspace{-0.03cm}0$ dominates (i.e., ROPT crossing first) for vehicles to the left.
%While there is more area of $\text{PET}\hspace{-0.05cm}>\hspace{-0.05cm}0$ (i.e. ROPT crossing first) when the car comes from the left, the condition $\text{PET}\hspace{-0.05cm}<\hspace{-0.05cm}\unit[-2]{\text{sec}}$ dominates (i.e. ROPT driving second) for vehicles to the right. 
The transition from positive to negative values is on average at $v_{f,2}\hspace{-0.05cm}=\hspace{-0.05cm}\unit[7]{\text{m/sec}}$ in the former and about $v_{f,2}=\unit[10]{\text{m/sec}}$ for the latter case. With smaller $a_{f,2}$, the sign change happens at greater $v_{f,2}$.
%The transition from positive to negative values is on average at $v_{f,2}=\unit[10]{\text{m/sec}}$ in the former and about $v_{f,2}=\unit[7]{\text{m/sec}}$ for the latter case. With smaller $a_{f,2}$, the sign change happens at greater $v_{f,2}$.

The reason can be well observed in the ego velocity course $v_1$. %An inferior ROPT is inclined to decelerate as far as 
A prioritized ROPT is inclined to accelerate with $v_{\text{up},1}$ as far as $\unit[17.5]{\text{m/sec}}$, because it assumes the obstacle to stop. If the encountered vehicle disobeys (e.g. $a_{f,2} \approx  \unit[3]{\text{m/sec}}$ and $v_{f,2}~\hspace{-0.1cm}\approx~\hspace{-0.1cm} \unit[7]{\text{m/sec}}$), ROPT will at some point halt and give way. %safely which creates the highest $j_1$. 
These situations are still safe but create the highest jerk (refer to Section \ref{sec:randomvar}). Vice versa once TP2 has priority, ROPT brakes frequently having $v_{\text{low},1}$ under $\unit[2.5]{\text{m/sec}}$. At the same time, accelerating back to desired $v_{d,1}$ takes more time with $\text{PET}=[\unit[-3]{\text{sec}},\unit[-20]{\text{sec}}]$. The velocity growth prediction of TP2 leads to cautious ego behavior. %ROPT also needs more time to accelerate back to $v_{d,f}$ with $\text{PET}$. 
Here, overtaking is still established in small initial other speeds $v_{f,2}$. For $v_{f,2}\rightarrow0$, the other car does not even interfere with the ego trajectory and $\text{PET}>\unit[10]{\text{sec}}$ holds.  

\subsection{Randomized Intersection Geometries} %Other Paths and Intersections
\label{sec:randomvar}

For our large-scale experiment, the unsignaled intersection is hereafter extended with statistical conditions for the simulation. Altogether, we randomize path geometries, agents' starting states and the priority compliance of the other participant. This enables us to discuss hazards and jerk caused or rather avoided by ROPT from car-to-car passings. 

\vspace{0.04cm}
\subsubsection{Simulation Setup}
While we reduce the driving limit to fixed $v_{\text{max}}\hspace{-0.07cm}=\hspace{-0.07cm}\unit[8.5]{m/\text{sec}}$, each individual run has different angles between the four roads and random lane widths. Moreover, the start and destination roads for the ego and other vehicle are stochastic.\footnote{A prerequisit is that the start roads are distinct and their paths intersect. The situation will therefore always correspond to the basic lateral types depicted in Figure \ref{fig:interaction}.} Both cars subsequently start with sampled velocities $v_{f,1}$ and $v_{f,2}$ from $\unit[3.0]{m/\text{sec}}$ until $\unit[8.5]{m/\text{sec}}$ having a set distance $d_{I,1} \hspace{-0.02cm}=\hspace{-0.02cm}d_{I,2}\hspace{-0.02cm}=\hspace{-0.02cm}\unit[45]{m}$ to the intersection edge. Here, the desired cruising velocity $v_{d,1}$ for ROPT is always equal to $v_{\text{max}}$ and $v_{d,2}$ of the other participant is also randomized including higher speeds $\leq \hspace{-0.01cm} \unit[10]{m/\text{sec}}$. 
We finally vary the compliancy of the second car. In $\unit[50]{\%}$ of the experiments, TP2 is inattentive and ignores ROPT as long as the center-to-center distance $d_{2}$ is above $\unit[10]{m}$. This results into particularly challenging situations if ROPT has priority and assumes the obstacle to yield.

Opposed to Section \ref{sec:analyticalvar}, the simulation applies a full multi-agent planning. ROPT steers as above the ego car. In addition, the other car is controlled dynamically: 
it posseses the same cost function to evaluate trajectories (see Section \ref{sec:trajeval}), but exploits a simpler mechanism for creating candidate trajectories (i.e., no full optimization from Section \ref{sec:trajgen}). In each time step, the other vehicle directly constructs $21$ differing acceleration/deceleration profiles and selects the best one among them.

After each run, the unfolded driving scene is evaluated. For computing risk levels, we introduce a measure termed two-dimensional headway $\text{TH}_{\text{2D}}$ which expands TH to account for lateral distances. With the help of constant velocity extrapolation, $\text{TH}_{\text{2D}}$ essentially indicates the  time when vehicle pairs will occupy or have occupied the same space. %within the next $\text{TH}_{\text{2D}}$ seconds. %2DTH is a spatio-temporal measure indicating if two vehicles will occupy of have occupied the same space within a $T$ seconds window, using constant velocity extrapolation.
In detail, $\text{TH}_{\text{2D}}$ is obtained by first taking the bounding box for each agent consisting of four corners at the current step $t$. In the following calculation, we enlarge this box with the length of $v_t \cdot \frac{T}{2}$ in both directions along their path (whereby $v_t$ represents the velocity of the participant at $t$ and $T$ is the extrapolation interval). This means that  the resulting shape can bent around corners and is not convex.
We lastly define $\text{TH}_{\text{2D}}$ to be the minimum $T$ once the shapes of two vehicles overlap. Alongside our $\text{TH}_{\text{2D}}$, the maximal value of the ego jerk course $r_{\text{max},1}$ is likewise gathered. As a reference, most passengers rate a jerk until $\unit[3]{m/\text{sec}^3}$ as 
acceptable \cite{powell2015} and in emergency trajectories jerks above $\unit[6]{m/\text{sec}^3}$ are common \cite{bagdadi2009}. To neglect comfort reduction because of high frequency motion, we filter beforehand peaks in $r_1$ with rolling means and a window factor of $W=\unit[0.5]{\text{sec}}$.\footnote{Note that the moving average filter is not used within ROPT and only reduces outliers from $r_{\text{max},1}$ for the evaluation.}%\footnote{Downstreaming a trajectory controller after ROPT will act as a low-pass filter in the same way.}

\begin{figure}[t!]
      \centering
      \resizebox{\linewidth}{!}{
      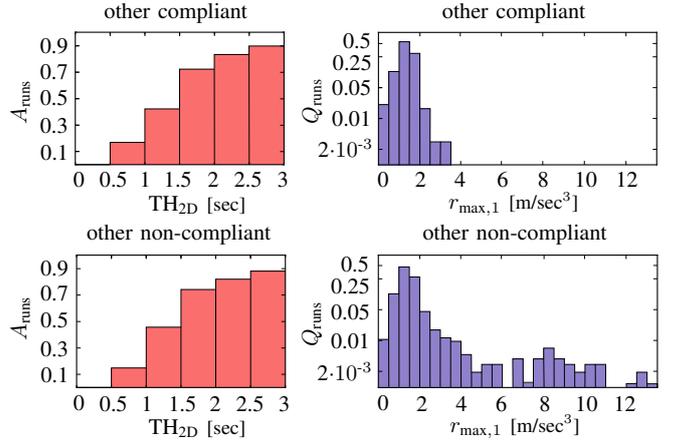} %
      \caption{Robustness of ROPT in diverse stochastic conditions during intersection crossings, e.g. priority violation of other car. Left: Cumulative histogram for two-dimensional headway. Right: Probability histogram for maximum ego jerk (note the log scale).} %(note the log scale).} %in log scale.
      %\caption{Post-Encroachment Time in 2-car intersection scenario with right vs. left priority. Top row: Other car approaching from right. Bottom row: Other car approaching from left. On an intersection, ROPT either drives more often after the other prioritized car ($\text{PET}<0$) or takes its precedence ($\text{PET}>0$).} %with normal comfort ranges %The comfort is never disregarded (low $j_1$)
      \label{fig:statistics}
\end{figure}

\vspace{0.04cm}
\subsubsection{Robustness Discussion}
More than 2000 simulations are executed involving the described settings. Figure \ref{fig:statistics} outlines the measured statistics for $\text{TH}_{\text{2D}}$ and $r_{\text{max},1}$. We initially focus on the ex-post risks.
The left side of Figure \ref{fig:statistics} renders cumulative distributions $A_{\text{runs}}$ for $\text{TH}_{\text{2D}}$. Regardless of TP2 following priority (top) or violating (bottom) priority, $\text{TH}_{\text{2D}}$ is larger than $\unit[1]{\text{sec}}$ in at least $\unit[85]{\%}$ of runs and $>\hspace{-0.04cm}\unit[0.5]{\text{sec}}$ in all runs.
In the cases when $\unit[0.5]{\text{sec}}\hspace{-0.02cm}<\hspace{-0.04cm}\text{TH}_{\text{2D}} \leq \unit[1]{\text{sec}}$, ROPT rightfully exerts its priority and the other vehicle crosses right behind. Decreasing values of $\text{TH}_{\text{2D}}$ are a consequence of the parametrization for TP2. It has higher escape rates $\tau_0$, which in effect lead to shorter prediction horizons and more aggressive planning. However, the main observation is that the trajectories of ROPT are always safe. ROPT must have compensated the incompliance of the other car and we thus look now more closely into the behavior of ROPT. % -- we will therefore look more closely to the ego vehicle in the next paragraph.

Probability distributions $Q_{\text{runs}}$ of the maximum jerk $r_{\text{max},1}$ encountered by ROPT are given in Figure \ref{fig:statistics} on the right.
If the other vehicle obeys right-of-way, $r_{\text{max},1}$ is below $\unit[2]{m/\text{sec}^3}$ in almost any situation (approximately $\unit[99]{\%}$), i.e. the ride feels comfortable. ROPT is robust against the intersection geometry or differences in taken starting and desired speeds from TP2. Rising $r_{\text{max},1}$ solely appear when TP2 has a counteracting behavior with non-compliancy to right-of-way. In such instances, ROPT has to compensate others' negligence by accepting larger $r_{\text{max},1}$. Usually, it reacts by either clearing the intersection earlier with accelerating away or by making a full brake to let TP2 pass in front. The latter can produce in $<\unit[1]{\%}$ highest $r_{\text{max},1}$ with up to $\unit[13]{m/\text{sec}^3}$. Nevertheless even for inattentive other participants, the jerk of ROPT is low to moderate for $\unit[90]{\%}$ of the situations and ROPT is able to smoothly adjust its behavior.

\vspace{0.4cm}
\section{Conclusion and Outlook}
\label{sec:outl}
In this work, we presented an optimization framework to predictively plan dynamic velocity curves under right-of-ways. ROPT considers risk (expected damage caused by collision and curvature), utility (distance travelled plus deviation to desired speed) and comfort (strength or frequency of behavior change) in one scalar cost function. The chosen parametric snake profile is composed of multiple ramps with an initial variable lag and smoothed transition points. Minimal and maximal values from engine and brake characteristics thereby maintain in ROPT realistic driving constraints.  %driving

%present
After geometrically determining the trajectory relationship between vehicle pairs, ROPT discounts the corresponding collision risk if other cars are longitudinally inferior (i.e., to the back). In lateral traffic situations, other participants are simultaneously assumed to prototypically decelerate when inferior (e.g. on the left) or accelerate when superior (e.g. on the right). %toafter a short-term constant velocity phase  
In this context, left-before-right settings can be easily applied in the same way. Furthermore, the fixed predictions are enhanced for ROPT by adhering to possible velocities in curves and to permitted road limits.
%, ROPT is hence superior to obstacles to the right and inferior inferiority applies to left and superiority to right relations.

Analytical experiments demonstrated that our method is able to effectively follow priority rules when necessary. While ROPT admits lower distances to following than leading cars for push reduction from tailgaiting, it drives in more instances before other vehicles when having precedence at intersections. Subsequently with path randomizations, we also proved that if the encountered obstacle is inattentive, ROPT avoids an accident while having good risk-comfort tradeoffs. Otherwise, the order of who goes first is safely kept and ROPT proactively manages intersection passings.

At this point, the path relations (e.g. front or back) are matched online with given regulatories (e.g. front-before-back). In large road junctions however, map data usually predefines the lane ranks %, e.g. turning main road or roundabouts. 
and other priority elements, such as traffic signs, should be incorporated as well. On this matter, the application of all-way stops shows to be promising, because they require to flexibly assign priority based on arrival time and might have superior pedestrian crossings. 
%Another direction would be four-way stops that dynamically assign priority based on who arrives first. 

Overall, ROPT is interaction-aware and could create cooperative behaviors in highway situations when permitting drivers on ramps to overtake or by prioritizing participants from faster lanes. In a next step, not only the interaction between pairs, but among all involved agents needs to be covered in the planning scheme. Rule deadlocks when cars on each incoming lane come together at an intersection are then solvable.

At last, real-time capability on a modern processor is not yet guaranteed with ROPT. %ROPT ran in the experiments with a time step of $\unit[200]{\text{msec}}$ on an Intel Core i5-6500 CPU and 8 \hspace{-0.1cm}GB RAM. 
For the application on a test car, it is possible to improve the computation time in two ways. On the one hand, the optimizer can leverage a numerical gradient from the explicit risk modeling. Parallely applied constructive heuristics (i.e., ramps are fine-tuned in succession to obtain one complete snake) enable here to get quicker out of local minima. On the other hand, a controller which executes the found ego trajectory for some timesteps supports ROPT to plan with lower update frequencies.

\vspace{0.3cm}

\section*{Acknowledgment}
\noindent This work has been partially supported by the European Unions Horizon 2020 project \textit{VI-DAS}, under the grant agreement number 690772. The authors would like to thank Fabian M\"uller for providing the damage model.
%\vspace{0.7cm}
\vfill



%\bibliographystyle{IEEEtran}
%\begin{thebibliography}{1}


%\end{thebibliography}

\bibliographystyle{IEEEtran}
%{\footnotesize \bibliography{bib.bib}}
\bibliography{bib}

\end{document}